\documentclass{article} 
\usepackage[T1]{fontenc}
\usepackage{hyperref}
\usepackage{amsmath}
\usepackage{amssymb}
\usepackage{mathtools}
\usepackage{amsthm}
\usepackage{algorithm}
\usepackage{algorithmic}
\usepackage{booktabs}
\usepackage{pifont}
\usepackage{graphicx}
\usepackage{wrapfig,lipsum}
\usepackage{microtype}
\usepackage{subcaption}
\usepackage[preprint]{neurips_2022}
\usepackage{times}
\usepackage{xcolor}

\title{Hierarchical Attention Encoder Decoder}

\author{
Asier Mujika \\
Department of Computer Science\\
ETH Zürich\\
Zürich, Switzerland \\
\texttt{asierm@inf.ethz.ch} \\
}

\let\cite\citep
\begin{document}

\maketitle

\begin{abstract}
Recent advances in large language models have shown that autoregressive modeling can generate complex and novel sequences that have many real-world applications. However, these models must generate outputs autoregressively, which becomes time-consuming when dealing with long sequences. Hierarchical autoregressive approaches that compress data have been proposed as a solution, but these methods still generate outputs at the original data frequency, resulting in slow and memory-intensive models. In this paper, we propose a model based on the Hierarchical Recurrent Encoder Decoder (HRED) architecture. This model independently encodes input sub-sequences without global context, processes these sequences using a lower-frequency model, and decodes outputs at the original data frequency. By interpreting the encoder as an implicitly defined embedding matrix and using sampled softmax estimation, we develop a training algorithm that can train the entire model without a high-frequency decoder, which is the most memory and compute-intensive part of hierarchical approaches. In a final, brief phase, we train the decoder to generate data at the original granularity. Our algorithm significantly reduces memory requirements for training autoregressive models and it also improves the total training wall-clock time. 
\end{abstract}

\section{Introduction}

Autoregressive modeling has been widely employed to model various types of sequential data, including language \cite{raffel2020exploring}, images \cite{ramesh2021zero}, and audio \cite{oord2016wavenet}. In recent years, autoregressive models have gained significant attention due to their impressive scaling properties \cite{kaplan2020scaling, openai2023gpt4, fedus2022switch}. By training on larger datasets, increasing model sizes, and utilizing prompt engineering \footnote{Prompt engineering refers to modifying the context of a general autoregressive model to achieve a concrete goal.}\cite{peng2023instruction}, these models can achieve outstanding results in tasks such as question answering\cite{raffel2020exploring}, summarization\cite{raffel2020exploring}, and image completion \cite{parmar2018image}, among others.

Transformers \cite{vaswani2017attention} serve as the primary model for large-scale autoregressive learning. However, these models must operate at the same frequency as the original data, resulting in high computational costs. In language modeling, this problem is partially mitigated through tokenization\cite{kudo2018sentencepiece}, where frequent co-occurring character sequences are replaced by a single token. Nevertheless, models aiming to learn dependencies at the sentence level must still handle the original input data frequency. In domains with even higher-frequency inputs, such as images or sound, this issue becomes a primary bottleneck. A large enough model capable of capturing the data distribution would be too slow and memory-intensive to operate at the original data's frequency. To put this in context, GPT-4, one of the most powerful language model trained to date, has a context window of 25,000 tokens, which would barely suffice to capture the individual pixels of a 92x92 RGB image in an autoregressive manner.

To address this issue, we revisit the Hierarchical Recurrent Encoder-Decoder (HRED) architecture \cite{sordoni2015hierarchical}, which separates the computationally intensive high-frequency data from the more complex low-frequency interactions. This model divides the input sequence into non-overlapping sub-sequences that are encoded independently of each other. The resulting lower-frequency vector stream is then processed by a potentially larger model, no longer bottlenecked by the original data's frequency. Finally, the output vectors are individually mapped back to the original data frequency. Although subsequent downsampling approaches have been proposed in the transformer literature, most of them feature an encoder or decoder that globally handles the entire original sequence, resulting in significant computational and memory cost. By encoding and decoding sub-sequences independently, the HRED model can circumvent these issues.

In this paper, we modify the HRED model to enhance its performance and propose a learning algorithm specifically designed for this architecture. Our main contributions include:

\begin{itemize}
    \item Analyzing the different components of the HRED to identify which parts contribute the most to model performance.
    \item Based on these insights, we propose a modification of the HRED, termed the Hierarchical Attention Encoder-Decoder (HAED) architecture, that considerably improves the performance of the original model. Our model replaces the recurrent neural networks in the encoder and main model by MLPs and transformers, respectively.
    \item Introducing a learning algorithm for the HAED, capable of learning with low-frequency targets directly, which significantly reduces compute time and consequently allows for larger models and greater amounts of training data.
\end{itemize}
\section{Related Work}

Our work takes inspiration from the Hierarchical Recurrent Encoder-Decoder (HRED) model \cite{sordoni2015hierarchical}, which was initially proposed to independently encode user-generated queries and decode new queries from the context created by all previous queries. Our approach diverges by incorporating Transformers and MLPs instead of the original RNNs, applying the method to character and pixel modeling, and proposing a new learning algorithm specifically designed for this architecture.

Numerous hierarchical approaches have been suggested in the recurrent network literature, mainly to tackle the vanishing gradients problem \cite{jaderberg2019human, chung2016hierarchical, koutnik2014clockwork, zhao2018hsa}. However, most of these approaches employ the same model for encoding and decoding, which prevents independent analysis of both mechanisms and the development of decoupled learning algorithms—two key contributions of our paper. Additionally, many of these studies attempt to learn the hierarchy itself \cite{chung2016hierarchical, zhao2018hsa}, a problem that we do not explore in this paper.

As the vanishing gradients problem is not as prominent in attention-based architectures as it is in RNNs, most attention-based hierarchical approaches concentrate on restricting the attention mask to attend only to sparse events in the sequence \cite{ho2019axial, parmar2018image, child2019generating, ren2021combiner}. While this reduces the computational costs of the attention mechanism, the model itself remains non-hierarchical and never reduces the input frequency. The Hourglass Transformer \cite{nawrot2021hierarchical} is the most similar approach to ours. It features an encoder, decoder, and main model akin to our approach and the HRED model. However, since the encoder has global context, it cannot be interpreted as an embedding matrix, making our learning algorithm inapplicable to it. Furthermore, as both the encoder and decoder deal with global sequences, they experience typical issues with long sequence processing, such as large sequential computation and gradient decay with recurrent models or poor compute/memory scaling with very long sequences of attention-based models. Other similar approaches aimed at reducing sequence length for transformer architectures can be found in \cite{dai2020funnel, nawrot2022efficient, clark2022canine, hawthorne2022general}.

A similar approach to our implicit embedding matrix learning algorithm was proposed by \cite{oord2018representation}, which aims to maximize the information between the current global context and future encoded sub-sequences to learn unsupervised low-frequency representations of input data. Similarly, CLIP \cite{radford2021learning} applies this technique to match images to captions. They demonstrated that this approach outperforms learning an autoregressive decoder of captions from images, which is unsurprising considering our implicit embedding matrix interpretation. Their algorithm can be seen as simultaneously learning an autoregressive decoder of captions given images and images given captions, which mostly likely improves the learning signal.

The idea of replacing explicit embedding or output matrices with CNNs or RNNs was also investigated in \cite{jozefowicz2016exploring} as a final step in language modeling to reduce the number of parameters since those matrices contained the majority of parameters in their experiments. Our work can be regarded as a generalization of this approach, where the output/embedding matrix is never actually materialized, allowing for training on open-vocabulary language modeling and image modeling.

Contemporary work to ours \cite{contemporary}, has shown that a very similar model to the HRED and HAED, but with transformers for all models, can achieve state-of-the-art performance on many long-term dependency benchmarks. Based on our analysis of the different components, this is probably not the optimal way to distribute the compute. Still, our training algorithm can also be applied to that model. 
\section{Method}
\label{sec:methods}

In this section we introduce the Hierarchical Attention Encoder-Decoder (HAED) architecture together with a learning algorithm specifically designed for this model, the Implicit Embedding Matrix (IEM) algorithm. The architecture is a modification of the Hierarchical Recurrent Encoder-Decoder (HRED) \cite{sordoni2015hierarchical} that replaces the Recurrent Neural Networks (RNN) in the encoder and the main model by feed-forward MLPs and Transformers, respectively. The learning algorithm enables learning the model without a decoder, which is the most computationally expensive part of the model.

To disentangle hierarchy learning from hierarchy exploitation, we employ hard-coded hierarchies and focus solely on the second part in this paper. For language modeling, we aggregate characters at the word level, and for image modeling, we take a fixed number of pixels at each hierarchical level.

\subsection{Hierarchical Attention Encoder-Decoders}

\begin{figure}[htb]
  \centering
    \includegraphics[width=0.6\textwidth]{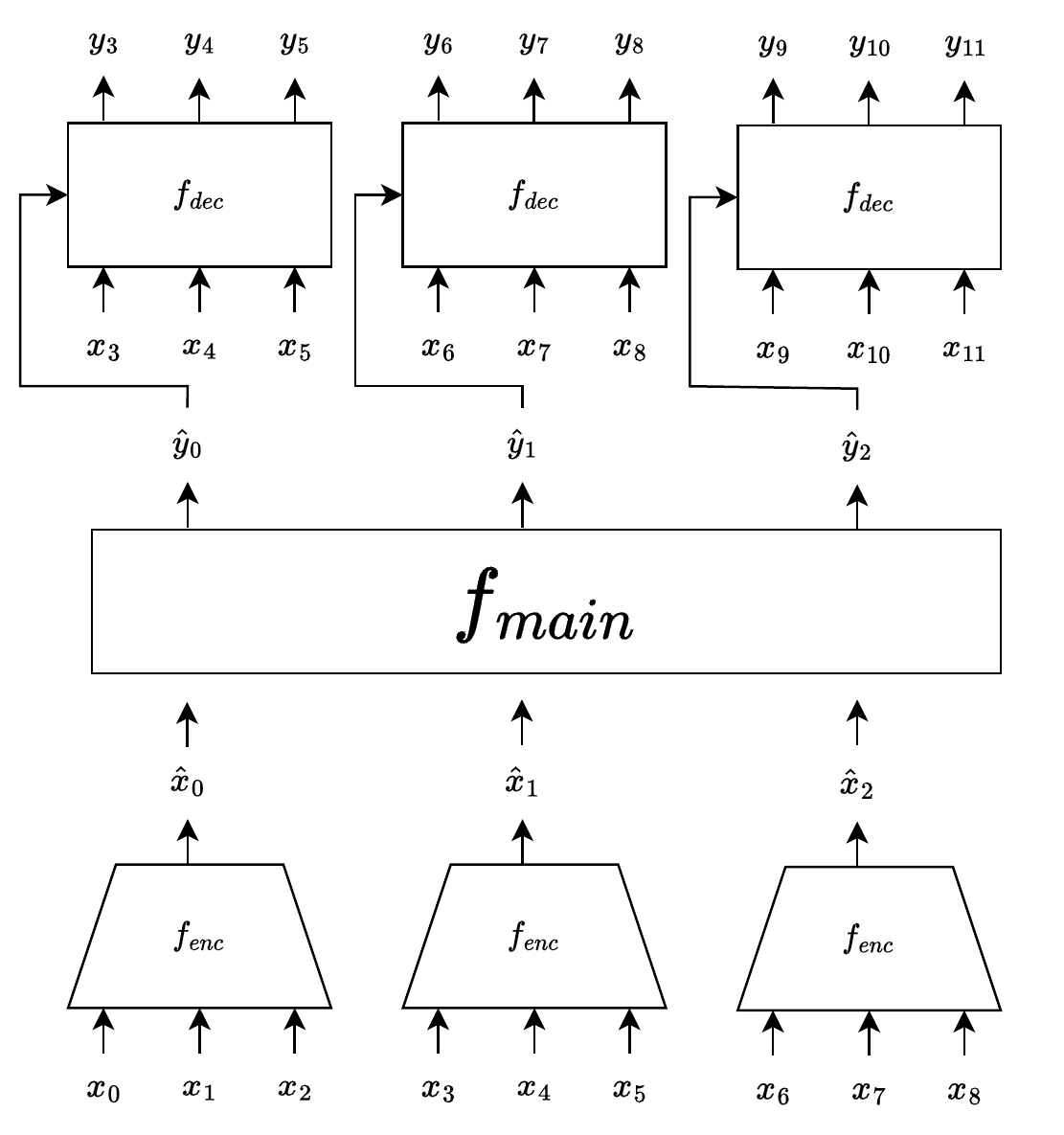}
    \caption{Diagram representing Equations \ref{eq:enc}, \ref{eq:main} and \ref{eq:dec}. In this example, the hierarchy aggregates the original inputs every $3$ steps. $f_{main}$ and $f_{dec}$ are causally masked, to preserve the autoregressive information flow.}
    \label{fig:haed}
\end{figure}

The HRED model divides the high-frequency input sequence into non-overlapping sub-sequences that are encoded individually and independently, see Equation \ref{eq:enc}. This gives rise to a new sequence that has a lower frequency than the original input. This new sequence is processed by the main model, to generate low-frequency predictions, see Equation \ref{eq:main}. Finally, given the corresponding output from the main model, the decoder autoregressively generates the outputs at the frequency of the original sequence, see Equation \ref{eq:dec}. For simplicity, we use a fixed hierarchy of $k$ steps, but translating these equations to input-dependent hierarchies would be straightforward. See Figure \ref{fig:haed} for a diagram of the model.

\begin{align}
    \hat{x}_i &= f_{enc}(x_{i \cdot k}, \ldots, x_{k - 1 + i \cdot k}) \label{eq:enc}\\
    \hat{y}_i &= f_{main}(\hat{x}_0, \dots, \hat{x}_i) \label{eq:main}\\
    y_{i + j \cdot k} &= f_{dec}(x_{j \cdot k}, \ldots, x_{i + j \cdot k} , \hat{y}_j) \label{eq:dec}
\end{align}

In the original HRED paper, the authors employ RNNs for all of these functions. Our Hierarchical Attention Encoder-Decoder (HAED) architecture introduces two changes to the HRED model. First, we replace the encoder model with a feed-forward MLP. As demonstrated in Section \ref{sec:experiments}, the impact of $f_{enc}$ on performance is minimal, and replacing it with an MLP significantly accelerates the encoding without any decrease in performance. Second, we utilize a more advanced Transformer architecture for $f_{main}$. This results in considerably better performance than the original RNN, without causing a major increase in compute or memory requirements, as this model operates at a lower frequency.


\subsection{Training on the embedding space}

As shown in the Experiments section, the decoder, which operates at a high frequency, significantly affects the final performance. Consequently, most of the computation is required for the decoder. To address this, we introduce the Implicit Embedding Matrix (IEM) algorithm, that trains the model without the decoder, making predictions directly in the embedding space defined by the encoder.

We make a key observation for developing our learning algorithm: the encoder can be viewed as an implicit embedding matrix. As each sub-sequence is processed individually and without context, the encoder serves as a fixed mapping from a subsequence to a vector. This way, the encoder effectively defines an embedding matrix, where each entry corresponds to a possible subsequence, and the value of the entry is the output of the encoder.

Another technique employed in our algorithm is commonly used in autoregressive modeling: using the transpose of the embedding matrix to compute individual token probabilities. Using the implicit embedding matrix interpretation, we express the probability assigned to the ground truth label as:

\begin{align}
    &logit_i = f_{enc}(x_{i \cdot k}, \ldots, x_{k - 1 + i \cdot k}) \cdot \hat{y}_i \\
    &p(x_{i \cdot k}, \ldots, x_{k - 1 + i \cdot k}|x_0, \ldots, x_{t-1}) = \frac{e^{logit_i}} {Z}
\end{align}

However, the softmax normalizing factor $Z$ cannot be computed, as the embedding matrix is potentially exponential in size. For instance, in our image experiments, this matrix has $2^{96}$ entries. Many approaches have been proposed for sampling and reweighting negative samples when the softmax matrix is too large to be dealt with explicitly. In this paper, we use the simplest approach of taking negative samples from the same batch to compute $Z$. While this is a biased estimate \cite{blanc2018adaptive}, it suffices to achieve good experimental results. We believe that better negative sampling techniques can considerably improve performance, but we leave that for future work.

As a final step to either sample or perform exact density estimation of the data, an actual decoder must be trained. However, as shown in Section \ref{sec:experiments}, this can be done much faster after training with our algorithm. 

\section{Experiments}
\label{sec:experiments}

In this section, we will first experimentally evaluate the different components of the HRED and experimentally justify the introduction of the HAED architecture and next, we will assess the IEM algorithm and precisely meassure the advantages it offers over end-to-end learning. 

We will evaluate relatively small models (around 10 million parameters) on relatively large datasets, character-level WikiText-103 \cite{merity2016pointer} and pixel-level \footnote{Each pixel is represented by $3$ tokens, each corresponding to a RGB channel} Imagenet-32 \cite{deng2009imagenet, van2016pixel}. We will do this in order to avoid data reuse and thus, ignore overfitting, thereby reducing potential confounding effects. To further reduce possible confounding factors, we will use a hard-coded hierarchy. For character-level modeling we will encode/decode once per word and in the pixel-level modeling, we will encode/decode every $4$ pixels, that is, every $12$ inputs.

In order to improve the readbility of this section but still ensure reproducibility, we move all the experimental details and hyperparameters to the Appendix. 

\subsection{Experimental evaluation of HAED model}
\label{sec:haed-exps}

As a first step towards our HAED model we replace the LSTM in the main model of the HRED by a transformer. 
To validate this decision, we train two HRED models, one with a LSTM as a main model and one with a Transformer. Both models have roughly the same number of parameters. Table \ref{table:HRED-attention} summarizes our results. As expected, transformers performed much better than LSTMs. Unsurprisingly, the HRED with a Transformer considerably outperforms the original model.

\begin{table}[htb]
  \centering
  \caption{We train a LSTM based HRED and another one where the main model LSTM is replaced by a Transformer. Both models train on the same amount on data and have roughly the same number of parameters. The results are reported in bits per input token. \vspace{1em}}
  \label{table:HRED-attention}
  \begin{tabular}{ccc}
    \hline
    \textbf{Dataset}  & HRED & HRED with Transformer \\
    \hline
    WikiText-103 & 1.52 & 1.37\\
    Imagenet32 & 4.13 & 4.05\\
    \hline
  \end{tabular}
\end{table}

\subsubsection{Encoder Model}
\label{sec:encoder-exps}

\begin{figure}[htb]
  \centering
  \begin{subfigure}[b]{0.4\textwidth}
    \includegraphics[width=\textwidth]{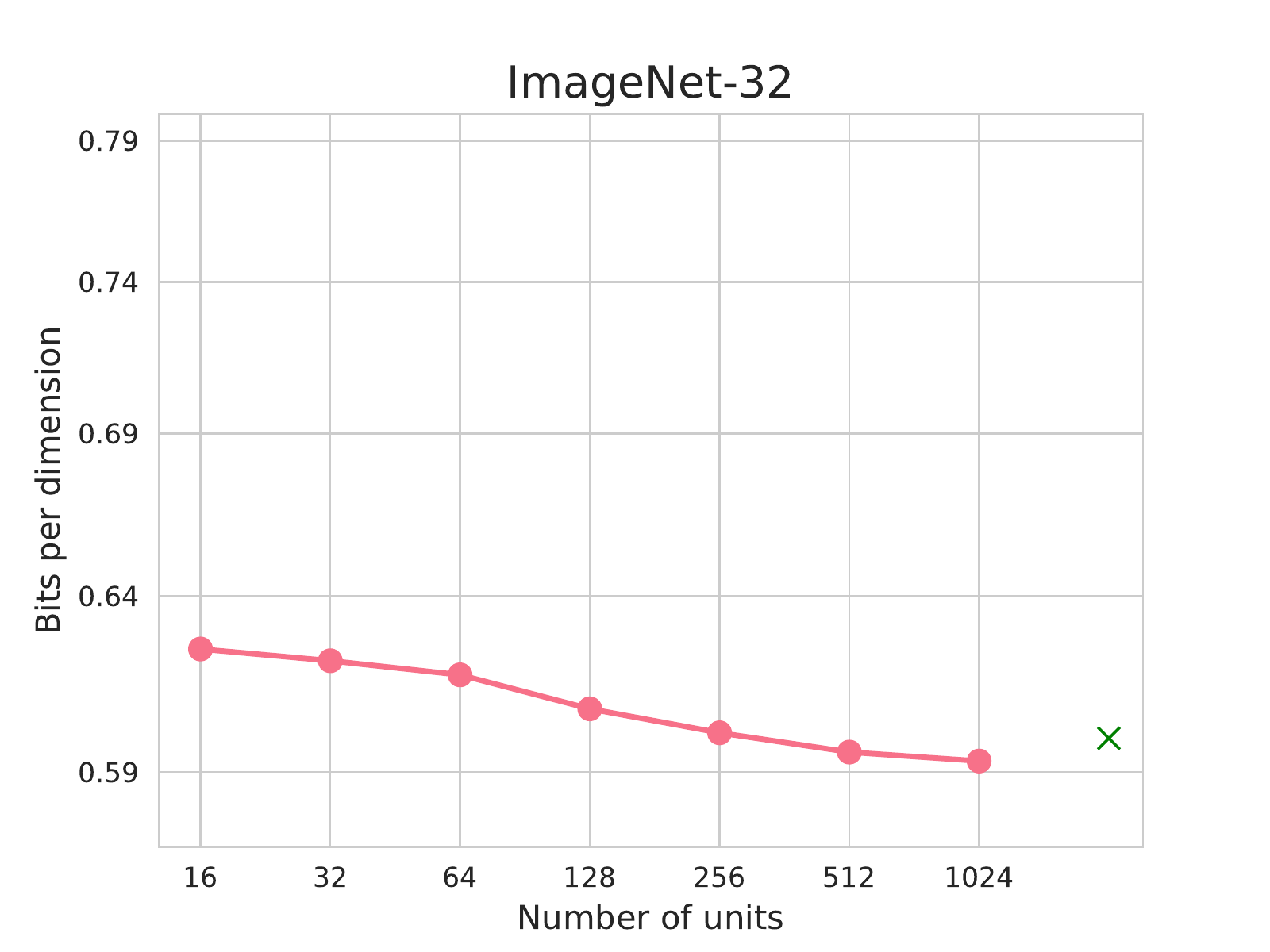}
    \caption{ImageNet-32 performance}
    \label{fig:subfig1}
  \end{subfigure}
  \hspace{0.05\textwidth}
  \begin{subfigure}[b]{0.4\textwidth}
    \includegraphics[width=\textwidth]{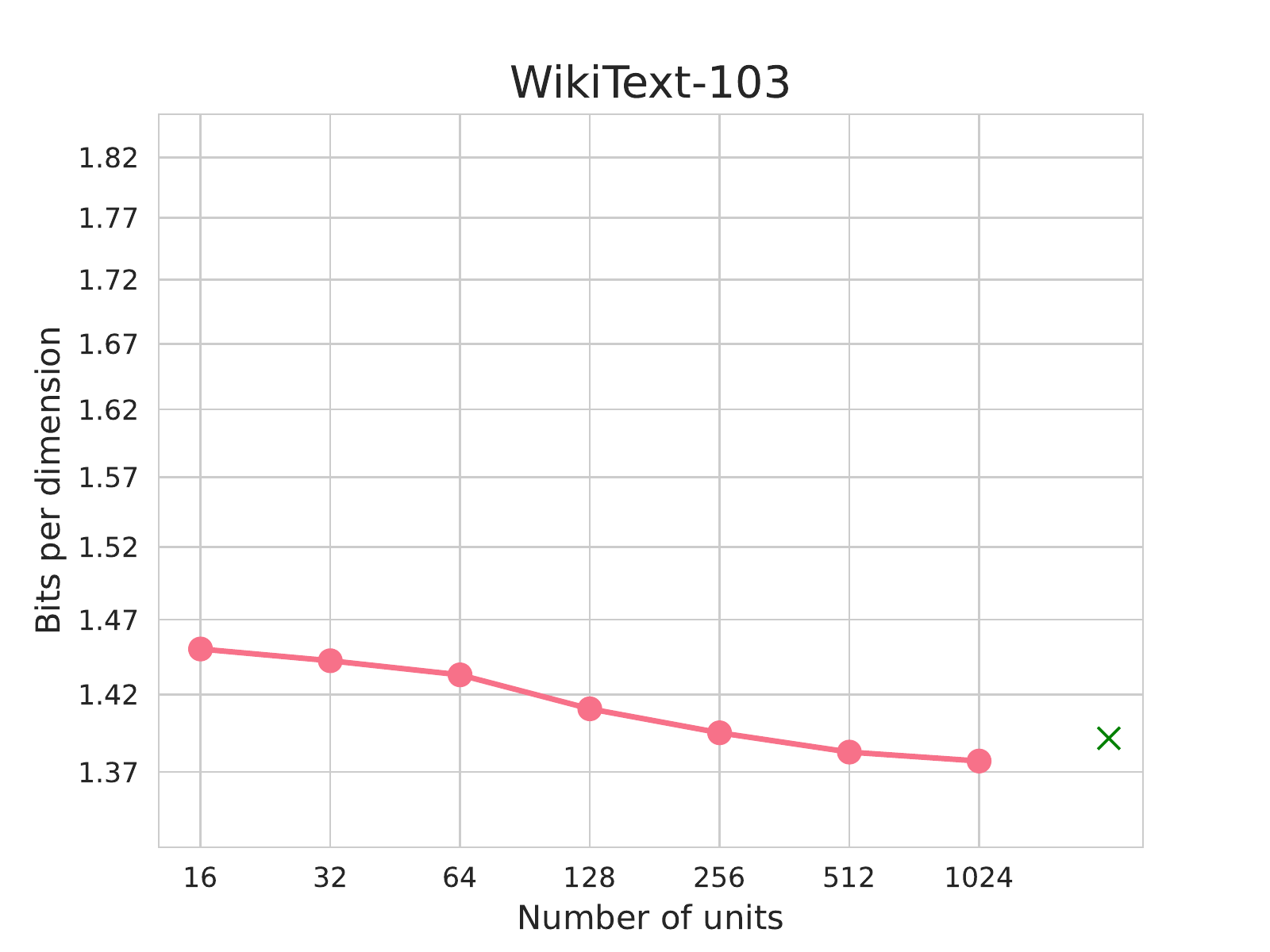}
    \caption{WikiText-103}
    \label{fig:subfig2}
  \end{subfigure}
  \caption{Model loss as a function of the number of units in the encoder. The "x" corresponds to the performance of the MLP model. The y-axes are set to the same range as in Figure \ref{fig:decoder}.}
  \label{fig:encoder}
\end{figure}

The encoder's context is limited, which implies that its computation should be straightforward and not gain much from increased model capacity. To confirm this intuition, we train the attention based HRED from the previous section with LSTM encoders that have varying numbers of units, keeping the FLOPS/parameters of the rest of the model fixed, and plot the resulting impact on the final model performance. Figure \ref{fig:encoder} illustrates our findings, indicating that once the model has enough capacity to encode the sequence correctly, there is no further benefit in making the encoder larger.

Having this knowledge, we experiment with a simple 2-layer MLP as our encoder. This significantly speeds up the encoder, as it requires fewer FLOPS, allows for greater parallelization than sequential models and requires much less memory. Compared to a LSTM of equal final performance, it is faster and uses less memory. Therefore, for the remaining experiments, we will use this encoder architecture. We will refer to the model with a MLP encoder and Transformer main model as Hierarchical Attention Encoder Decoder (HAED).

\subsubsection{Decoder Model}
\label{sec:decoder-exps}

\begin{figure}[htb]
  \centering
  \begin{subfigure}[b]{0.3\textwidth}
    \includegraphics[width=\textwidth]{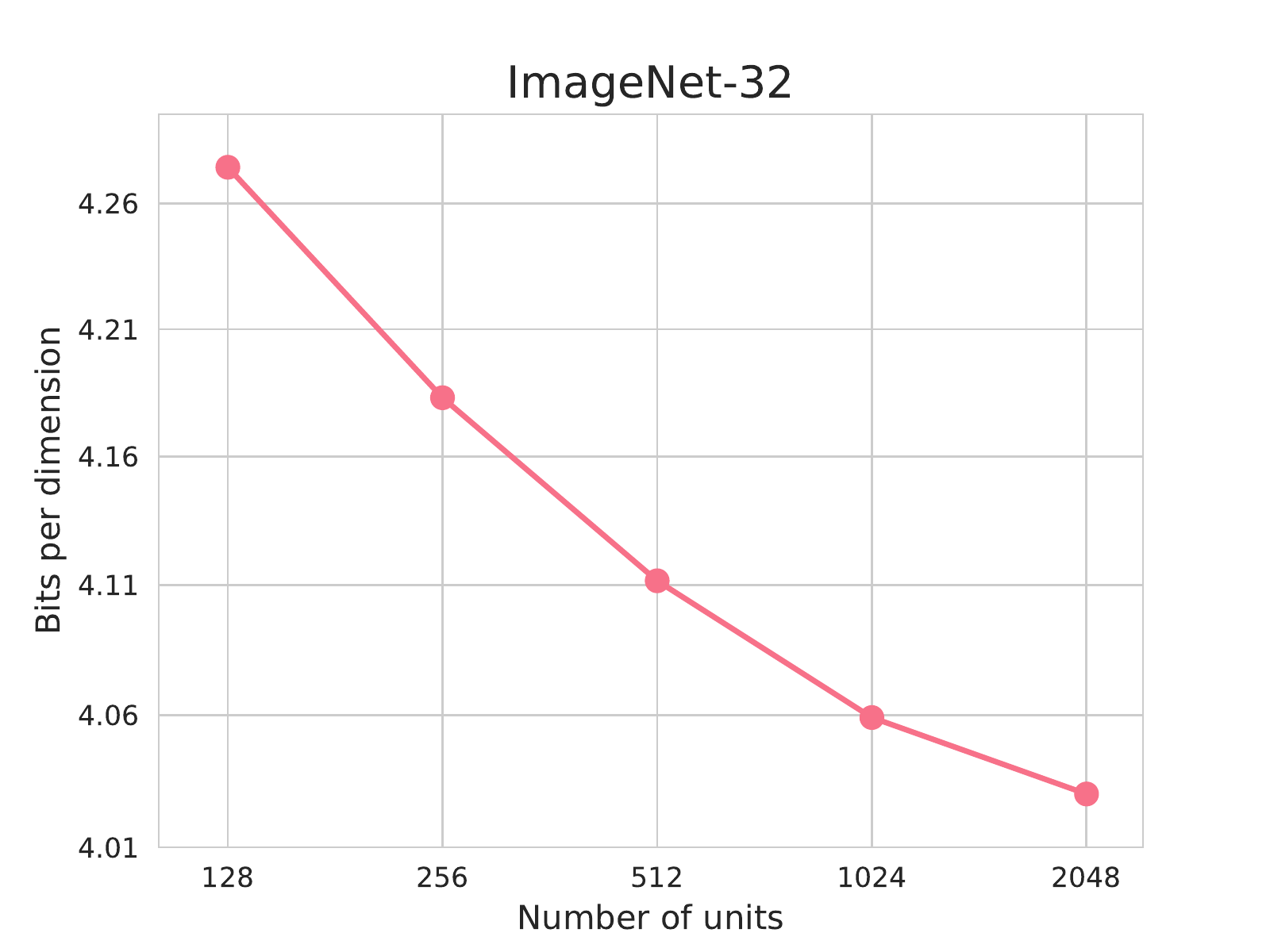}
    \caption{ImageNet-32 performance}
    \label{fig:subfig1}
  \end{subfigure}
  \hspace{0.02\textwidth}
  \begin{subfigure}[b]{0.3\textwidth}
    \includegraphics[width=\textwidth]{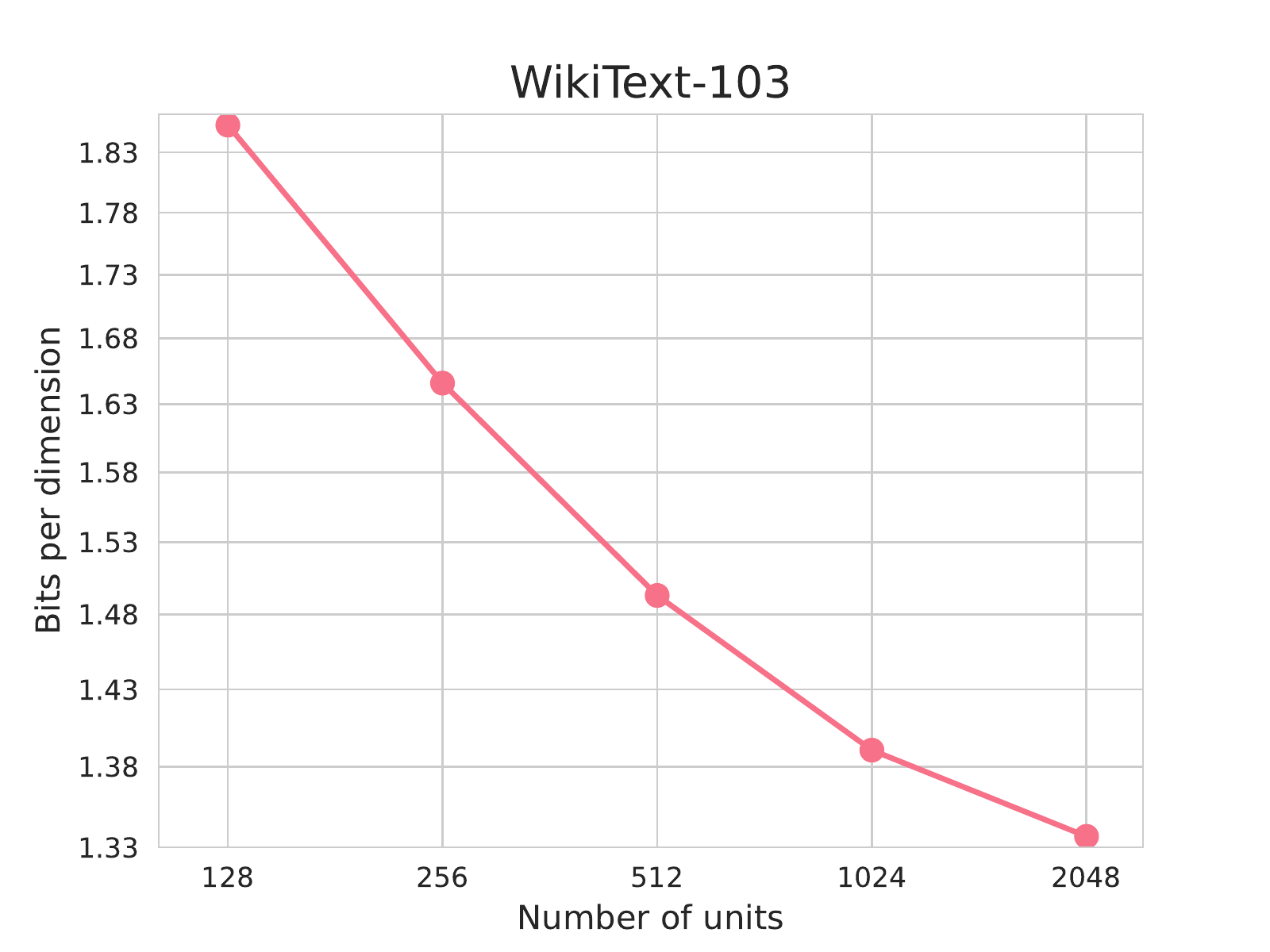}
    \caption{WikiText-103 performance}
    \label{fig:subfig2}
  \end{subfigure}
  \hspace{0.02\textwidth}
  \begin{subfigure}[b]{0.3\textwidth}
    \includegraphics[width=\textwidth]{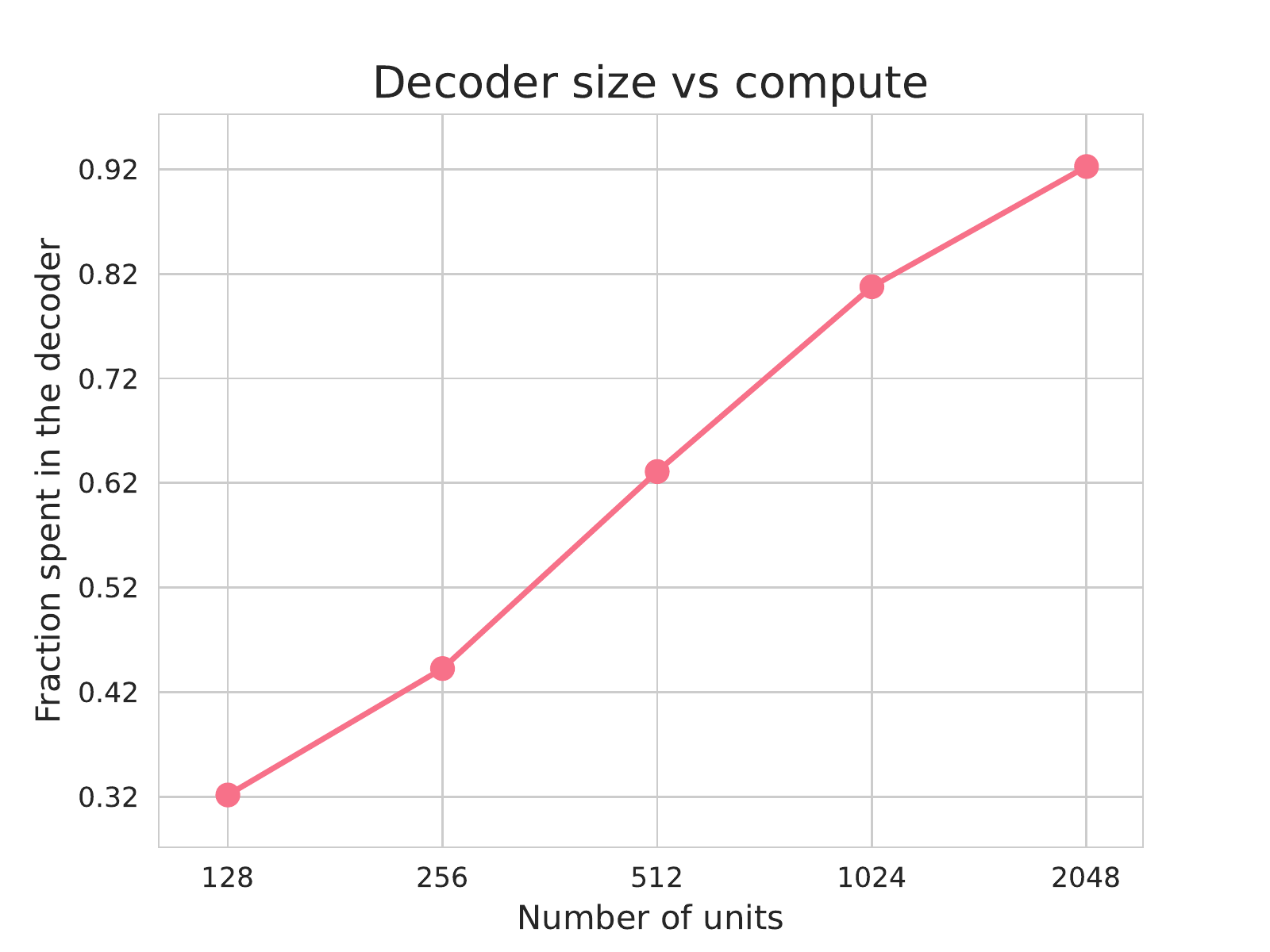}
    \caption{Run-time}
    \label{fig:subfig2}
  \end{subfigure}
  \caption{Model loss and percent of time spent in the decoder as a function of the number of units in it.}
  \label{fig:decoder}
\end{figure}

As with the encoder model, we want to measure the impact of increasing the decoder capacity on the overall model performance. Figure \ref{fig:decoder} shows the results of increasing the decoder capacity. In contrast to the encoder, we observe a continous improvement on the loss as we increase the decoder size. However, this comes at a great computational cost as the decoder is ran at a much higher frequency than the main model. It can be seen that the decoder compute time quickly becomes the main bottleneck for training the model.

\subsection{Decoupling}

\begin{figure}[htb]
  \centering
  \begin{subfigure}[b]{0.4\textwidth}
    \includegraphics[width=\textwidth]{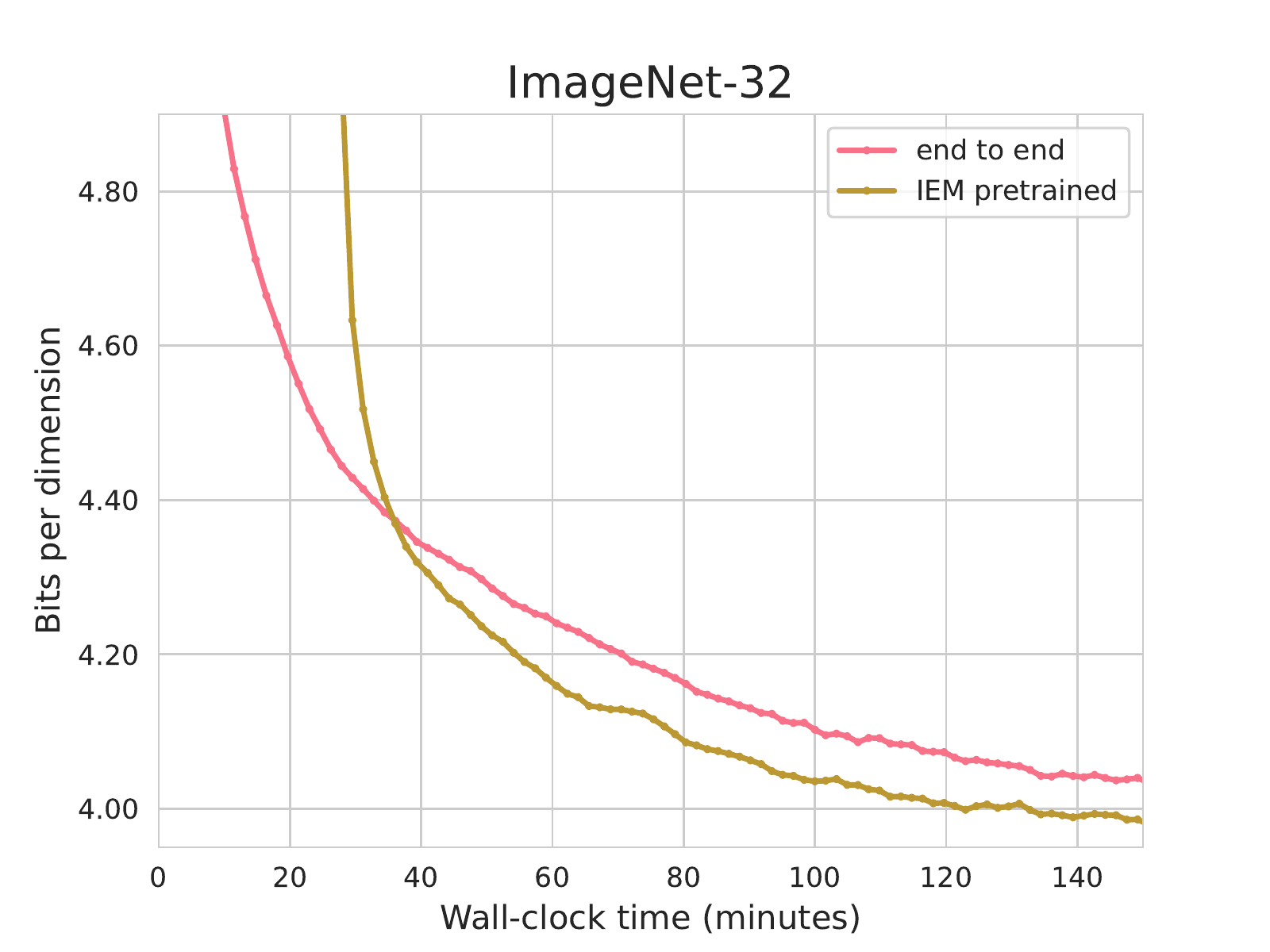}
    \caption{Imagenet-32 performance}
    \label{fig:subfig1}
  \end{subfigure}
  \hspace{0.05\textwidth}
  \begin{subfigure}[b]{0.4\textwidth}
    \includegraphics[width=\textwidth]{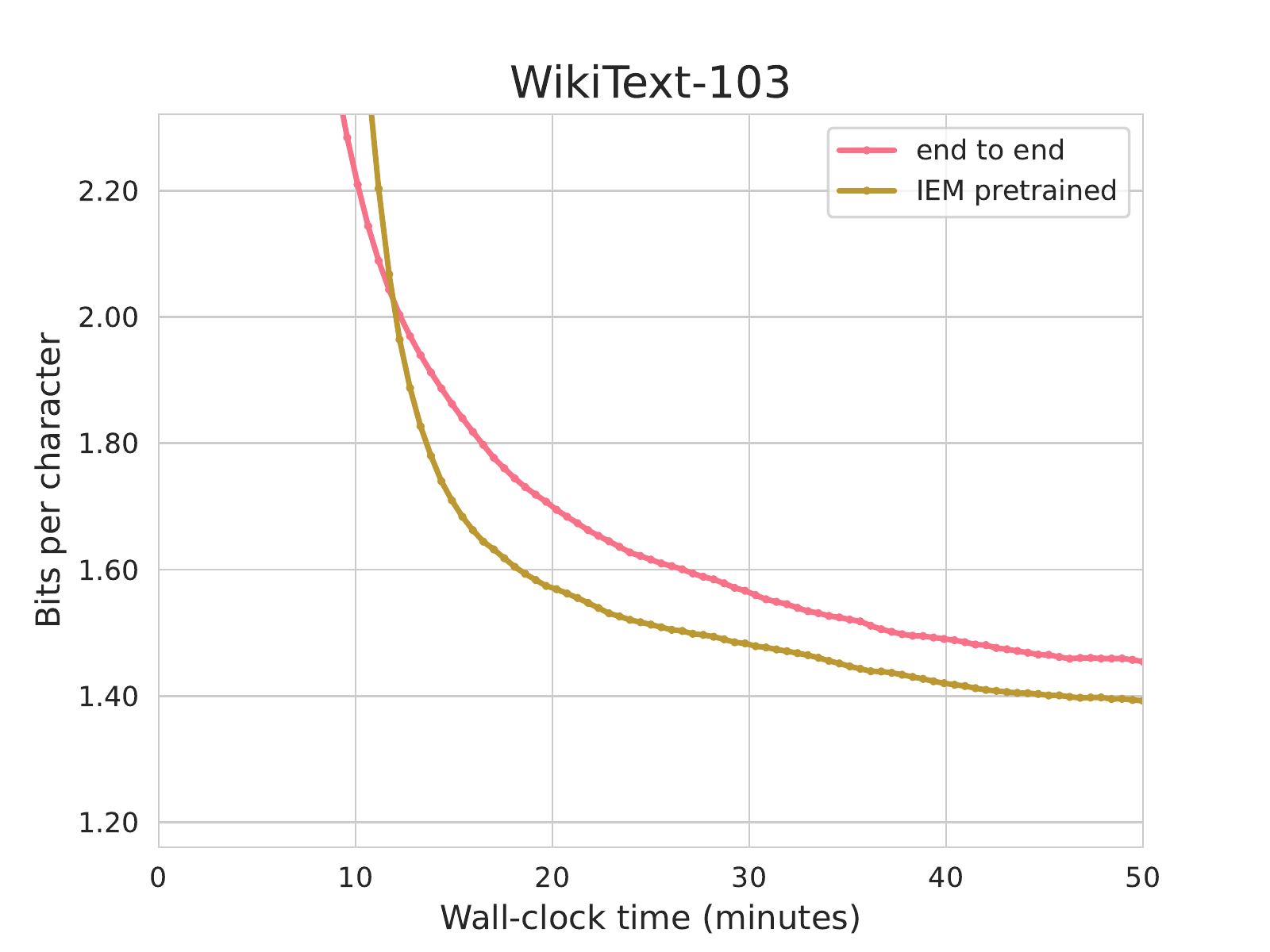}
    \caption{WikiText-103 performance}
    \label{fig:subfig2}
  \end{subfigure}
  \caption{Model loss as a function of total wall-clock time, including pretraining.}
  \label{fig:IEM}
\end{figure}

As illustrated in Figure \ref{fig:decoder}, the majority of the training time is spent on training the decoder in larger models. Therefore, we aim to evaluate the benefits of pre-training with the IEM algorithm, allowing us to avoid using a decoder during most of the training process. In order to assess the advantages provided by our algorithm, we train an encoder and the main model using our auxiliary loss on half of the data from each dataset. Then, we fine-tune the model end-to-end with the decoder on the other half of the data. For comparison, we train a model end-to-end for the same amount of time as a baseline. Figure \ref{fig:IEM} demonstrates that our auxiliary loss improves the performance of the model, even when pre-training time is taken into account.

\section{Future Work}
\label{sec:fut_work}

Our proposed approach offers many promising avenues for future work.

Firstly, we have only explored a single hierarchical level. While this has demonstrated significant advantages as shown in the sections above, adding more hierarchical levels could further improve the performance of the model.

Secondly, the way negative samples are selected is probably suboptimal. This selection method has been shown to have a considerable impact on the performance of models based on the sampled softmax technique. In fact, the unigram distribution (the one we are sampling from in our experiments) tends to perform much worse than other context-dependent distributions. We believe that with carefully crafted negative samples, the pretraining loss will become even more aligned with the true loss and further improve performance.

Another avenue for future work could involve learning the hierarchies instead of hard-coding them. While there is a large body of knowledge addressing this question \cite{chung2016hierarchical, graves2016adaptive, nawrot2022efficient}, we have not attempted to combine any of those techniques with our model for the sake of simplicity. However, we believe that exploring this direction presents a great opportunity for future work.

Lastly, a final direction of research that we find particularly exciting is sampling in the embedding space. Although learning in the embedding space with our algorithm greatly improves training, the model still needs to sample one data-point at a time in the high-frequency domain for inference. We believe our algorithm can be adapted to also sample in the low-frequency domain, which can then be decoded to the high-frequency domain in parallel. This would enable efficient autoregressive sampling in high-frequency domains like images or audio.

\section{Limitations}

While our approach and experiments have demonstrated promising results, there are several limitations worth noting. 

Firstly, our study has been conducted using relatively small models, especially in comparison to the multi billion parameter models typically utilized to achieve state-of-the-art results. This size discrepancy raises questions about the potential scalability of our approach. In the context of very large models, it is conceivable that the decoder's capacity may at some point cease to contribute to overall performance improvement. Under such circumstances, our Implicit Embedding Matrix (IEM) algorithm might lose its importance, as the main model would then become the primary computational bottleneck. However, it is in these exact situations that the Hierarchical Attention Encoder-Decoder (HAED) architecture could prove particularly valuable. This is due to its ability to operate at a lower frequency, thus reducing computational demands through architectural efficiency, as opposed to relying on the learning algorithm.

Another potential limitation of our work lies in the use of hard-coded hierarchies in our experiments. Although numerous techniques for learning hierarchies exist, none are flawless or demonstrate consistency on par with hard-coded hierarchies. It remains possible that our architecture might be particularly sensitive to errors within learned hierarchies. Therefore, the exploration of how our approach interacts with hierarchy learning methods is an interesting avenue for future research. 

Despite these limitations, we believe that our proposed method opens up new horizons for improving the efficiency and effectiveness of large-scale autoregressive models.
\section{Conclusion}

We have presented a thorough experimental study of the different components of the Hierarchical Recurrent Encoder Decoder (HRED), which we have used to propose an improved model which we term the Hierarchical Attention Encoder Decoder (HAED).

Apart from that, we have also presented a learning algorithm that learns directly in the embedding space, rather than getting the learning signal from the high-frequency outputs, such as individual pixels. We show that by training with our algorithm, we can improve the performance and wall-clock training time of our model. 

We believe that both our algorithm and model open many exciting possibilities for the use of large attention based models for autoregressive generation in domains where it wasn't previously possible. Given the great scaling properties that these models have shown in language modeling, their applications for audio or image generation present exciting possibilities for the future.

\section{Aknowledgement and Disclosure of Funding}

I would like to thank Angelika Steger, Felix Weissenberger and Filippo Graziano for fruitful discussions in early iterations of this work. I was supported by grant no. CRSII5\_173721 of the Swiss National Science Foundation.


\bibliographystyle{unsrtnat}
\bibliography{bibliography}

\newpage

\appendix
\section{Experimental details of the HAED model}\label{app:experimental-details-HAED}

For these experiments we use the Adam optimizer \cite{kingma2014adam} with decoupled weight decay \cite{loshchilov2017decoupled} and always clip the gradient norm to be at most $0.01$. We use a learning rate of $0.002$ for the encoder/decoder and $0.00035$ for the main transformer model.  We decay the learning rate to $0.05$ of its initial value over training following a cosine schedule and use $2000$ steps of linear warm-up for the transformer based experiments. We train all models for a single epoch. We use an embedding size of $10$ for the high-frequency tokens. While this is very small, we didn't observe any advantage with larger embedding sizes. This could be because WikiText uses only characters and pixel-based tasks are actually encoding a real value (even though we treat it as a discrete value for autoregressive learning). For character-level modeling, we use words to split the original sequence. In order to avoid inefficiencies in our implementation when dealing with long words, we split words that are longer than $12$ character long. For images, we split the image into subsequences of $12$ tokens. We sample batches of $32$ full images in ImageNet and of $256$ words in WikiText.

In Section \ref{sec:haed-exps} we use the same encoder and decoder model, a $1024$ unit LSTM. For the transformer, we use a small transformer, see Table \ref{table:small-transformer} for the exact configuration.
For the LSTM we use a $1500$ unit LSTM, which has a similar number of parameters if we count only the recurrent ones. If we would take the LSTM input matrices into account, we would need to carefully tune input size to optimize performance. While only counting the recurrent parameters overeestimates the performance of the LSTM, as it has more parameters than the transformer, the former still performs considerably better. We restrict the output of the LSTM to be of the same dimensionality as the transformer, such that the decoder still has the same number of parameters. We also found that the main LSTM could get unstable sometimes and we capped the input gate to be at most $1$ minus the forget gate, which has been shown to have no major impact on performance, but offers more stability as this ensures the cell state stays between $-1$ and $1$ \cite{melis2017state}. We used a learning rate of $0.002$ for the main model LSTM which was found by grid search.

\begin{table}[h]
  \centering
  \caption{The small Transformer architecture we use for most of our experiments.}
  \label{table:small-transformer}
\begin{tabular}{|c|c|}
  \hline
  Number of layers & $6$\\
  \hline
  Model dimensionality & $356$\\
  \hline
  Feed-forward dimensionality & $1424$\\
  \hline
  Number of heads & $8$\\
  \hline
  Dimensionality of heads & $32$\\
  \hline
  Positional Embeddings & Absolute learned embeddings \\
  \hline
  Positional Embedding size & $100$ \\
  \hline
\end{tabular}
\end{table}

In Section \ref{sec:encoder-exps}, we use the decoder and main transformer model as in the previous paragraph and only vary the number of units in the encoder. To always have the same input size to the main model, we pad the encoder output with $0$'s when the encoder output is smaller than the main models input and only take the first $356$ units from the decoder when it is larger. For the MLP, we concatenate the $12$ inputs into a vector and use the following 2 layer MLP: $(12 \cdot 10) \rightarrow 256 \rightarrow ReLU \rightarrow 256 \rightarrow ReLU$. 

In Section \ref{sec:decoder-exps}, we use the same transformer as before, see Table \ref{table:small-transformer}. For the encoder we use the 2-layer MLP we described before. For the decoder, we use a standard LSTM with varying number of units.

\section{Experimental details of the IEM algorithm}\label{app:experimental-details-IEM}

In this Section we use almost the same setup to the previous one. The only differences are that we use a transformer with $12$ attention heads instead of $8$, and a larger decoder with $2000$ units. As negative samples we use the targets in the same batch and and also sample $3$ more batches that are only used for negative samples.

\end{document}